\documentclass[lettersize,journal]{IEEEtran}

\usepackage{soul}
\usepackage{amsmath,amsfonts}
\usepackage{algorithmic}
\usepackage{algorithm}
\usepackage{array}
\usepackage[caption=false,font=normalsize,labelfont=sf,textfont=sf]{subfig}

\usepackage{textcomp}
\usepackage{stfloats}
\usepackage{url}
\usepackage{verbatim}
\usepackage{graphicx}
\usepackage{cite}
\usepackage{color}
\usepackage[table]{xcolor}
\usepackage{multirow}
\makeatletter
\newcount\SOUL@minus %% <- without this line the page number would be reset to 0
\makeatother

\begin{document}

\title{GaitPoint+: A Gait Recognition Network Incorporating Point Cloud Analysis and Recycling}
%\title{GaitPoint with Recycling Maxpooling Module: A Gait Recognition Network based on Point Cloud Analysis}

\author{Huantao Ren, Jiajing Chen, and~Senem Velipasalar~\IEEEmembership{Senior Member,~IEEE}
        % <-this % stops a space
\thanks{Authors are with the Department of Electrical Engineering and Computer Science, Syracuse University, Syracuse NY, 13244 USA e-mail: \{hren11,jchen152,svelipas@syr.edu\}. This work was supported in part by the National Science Foundation under Grant 1816732.}% <-this % stops a space
%\thanks{Manuscript received July 10, 2021; revised September 26, 2021.}
%\thanks{(*) Authors contributed equally.}
}
% The paper headers
% \markboth{Journal of \LaTeX\ Class Files,~Vol.~14, No.~8, August~2021}%
% {Shell \MakeLowercase{\textit{et al.}}: A Sample Article Using IEEEtran.cls for IEEE Journals}

%\IEEEpubid{0000--0000/00\$00.00~\copyright~2021 IEEE}
% Remember, if you use this you must call \IEEEpubidadjcol in the second
% column for its text to clear the IEEEpubid mark.

\maketitle

\begin{abstract}
Gait is a behavioral biometric modality that can be used to recognize individuals by the way they walk from a far distance. Most existing gait recognition approaches rely on either silhouettes or skeletons, while their joint use is underexplored.  Features from silhouettes and skeletons can provide complementary information for more robust recognition against appearance changes or pose estimation errors. To exploit the benefits of both silhouette and skeleton features, we propose a new gait recognition network, referred to as the GaitPoint+. Our approach models skeleton key points as a 3D point cloud, and employs a computational complexity-conscious 3D point processing approach to extract skeleton features, which are then combined with silhouette features for improved accuracy. Since silhouette- or CNN-based methods already require considerable amount of computational resources, it is preferable that the key point learning module is faster and more lightweight. We present a detailed analysis of the utilization of every human key point after the use of traditional max-pooling, and show that while elbow and ankle points are used most commonly, many useful points are discarded by max-pooling. Thus, we present a method to recycle some of the discarded points by a Recycling Max-Pooling module, during processing of skeleton point clouds, and achieve further performance improvement. We provide a comprehensive set of experimental results showing that (i) incorporating skeleton features obtained by a point-based 3D point cloud processing approach boosts the performance of three different state-of-the-art silhouette- and CNN-based baselines; (ii) recycling the discarded points increases the accuracy further; (iii) improvement margin is higher for more difficult scenarios involving appearance changes, such as carrying a bag or wearing heavier clothing. Ablation studies are also provided to show the effectiveness and contribution of different components of our approach. 
\end{abstract}


\section{Introduction}
\IEEEPARstart{G}{ait} is a behavioral
%and physical 
biometric modality that allows identifying individuals based on the way they walk. Other biometrics, such as iris~\cite{wildes1997iris}, fingerprint~\cite{hrechak1990automated} and face~\cite{sepas2020face} recognition, require subjects to be close to the acquisition system(s) and/or subjects' cooperation. In contrast, gait is an unobtrusive biometric that allows recognizing a person from a far distance without direct contact. Hence, gait recognition can be widely employed in many different application areas, including security and video surveillance. Yet, gait recognition has its own challenges. Different factors, including different camera angles and variances in appearances of individuals can affect performance of gait recognition. Heavy clothing, carrying items, such as a backpack, change appearance of people. Moreover, different camera viewpoints~\cite{wu2016comprehensive} and occlusions by an object or a part of an individual’s own body in certain camera views~\cite{uddin2019spatio} can also cause appearance variations. 
%many different exterior factors bring significant challenges to gait recognition, such as variations in appearance of individuals. Wearing outerwear, such as a coat, or carrying a bag can change the appearance of individuals. 

Gait recognition approaches can be broadly classified into two categories: appearance-based and model-based. Appearance-based methods operate on silhouettes to generate gait features. They either use the gait images directly or compress gait information into a single Gait Energy Image (GEI)\cite{wu2016comprehensive, han2005individual,rida2016gait} or Gait Entropy Image (GEnI)\cite{bashir2009gait}. Although a sequence of silhouettes contains useful gait features, such as gait cycle time, stride length, leg length and speed, silhouettes are susceptible to changes in appearance, such as different clothing and shoe types, hairstyle and scale. Model-based methods, on the other hand, mainly rely on the features of an articulated human model, such as a skeleton, and these features are less sensitive to appearance changes. However, it is hard to accurately locate and track a human body, and reliably obtain human pose and skeletons. Moreover, an individual’s body shape or silhouette still provide useful information for gait recognition, and thus, they should not be entirely discarded. Skeleton features and silhouette data can provide complementary gait information. 

Motivated by the above discussion, and inspired by the success of point-based methods in 3D point cloud processing, we propose a novel and general framework, referred to as GaitPoint+, for gait recognition. Our proposed method takes advantage of the benefits of both appearance-based and model-based methods by obtaining and combining both silhouette features and skeleton features. We adopt PointNet~\cite{qi2017pointnet} as an auxiliary module to extract features from skeleton key points, and replace the traditional max-pooling operation used in PointNet with a Recycling Max-Pooling (RMP) module~\cite{chen2022discard}. As will be detailed below, most existing point-based approaches employ the max-pooling operation as a symmetric function to obtain permutation-invariant features from 3D point clouds. Yet, this causes many potentially useful points and their features to be discarded~\cite{chen2022discard}. We use a RMP module to recycle these points for gait recognition. In addition, we employ an appearance-based method, which could be any image- and convolution-based gait recognition method, to extract silhouettes features. We then combine silhouette and skeleton features to obtain a more exhaustive and richer set of features for gait recognition. The main contributions of this work include the following:
%namely GaitSet, GaitPart and GaitGL,
%since most of current point-based approaches employ max-pooling operation as a symmetric function to obtain permutation-invariant features from 3D point cloud, which causes some potentially useful points to be discarded as well, therefore, we adopt Recycling Max-Pooling (RMP) module~\cite{chen2022discard} to recycle these discarded points. Meanwhile, we use a appearance-based method, namely GaitSet, GaitPart and GaitGL, to extract silhouettes feature, and then combine silhouette and skeleton features to provide a more exhaustive and richer feature set for gait recognition. The main contributions of this work include the following:
\begin{itemize}%[leftmargin=0.3cm]
\item We model sequences of skeleton key points as a 3D point cloud and employ a 3D point-based point cloud processing approach, instead of graph convolutional networks (GCN), to extract features from these key points.
\item We perform a detailed analysis of the utilization of every human key point and show that while elbow and ankle points are used most commonly, many useful points are discarded by traditional max-pooling. We incorporate a Recycling Max-Pooling module and a refinement loss to recycle some of these discarded points during skeleton point processing, and achieve increased point utilization and further performance improvement.
\item Since RMP is only employed during training, the performance improvement comes with only a slight increase in the training time without affecting the inference time.
%For promoting gait recognition, we employ RMP module in point-based method to improve representation in skeleton sequences via increasing joints utilization. 
%, and (iii) the confidence score. 
%
\item We provide experimental results and comparisons with three state-of-the-art (SOTA) baselines on the CASIA-B dataset~\cite{yu2006framework} showing that when we treat skeleton points as 3D point clouds, and employ the point cloud analysis model as an auxiliary module to extract the features, it consistently improves the performance of all three purely silhouette- and CNN-based methods. Moreover, recycling the discarded points increases the accuracy further. The improvement margin is higher for more difficult scenarios involving appearance changes, such as carrying a bag or wearing heavier clothing. 
\item We provide extensive ablation studies analyzing the effects of different point cloud processing methods, and the recycling max-pooling module. We show that our chosen point cloud analysis method is computationally more efficient than others.
\end{itemize}
%
%
%showing that (i) incorporating skeleton features obtained by a point-based 3D point cloud processing approach boosts the performance of three different state-of-the-art silhouette- and CNN-based baselines; (ii) recycling the discarded points increases the accuracy further; (iii) improvement margin is higher for more difficult scenarios involving appearance changes, such as carrying a bag or wearing heavier clothing. Ablation studies are also provided to show the effectiveness and contribution of different components of our approach. % Our proposed approach improves the performance of all the silhouette- and CNN-based baselines with varying degrees for the more challenging scenarios of carrying a bag and wearing a coat. This shows the generalizability of our approach. It has also been observed that the more different the contours of the query and gallery images are, the higher the achieved improvement is. We have also shown that 
% features from the point cloud analysis model not only alleviate the problem of CNNs
% overfitting to person contours but also increase the robustness to different view angles.
% This is one of the first works that analyses the utilization of different skeleton key points for gait recognition. We have shown that the contribution of certain points is larger than others. We plan to explore the effects of different key points further as our future work.
Compared to our conference paper~\cite{chen2022gaitpoint}, this work is improved and extended in multiple ways: (i) instead of using traditional max-pooling, we incorporate the recycling max-pooling (RMP) module and a refinement loss, when analyzing 3D point clouds formed from skeleton key points, and achieve better performance; (ii) we analyze the utilization of each human key point and explore the impact of different input key points on the results; (iii) we study and experiment with different point-based methods to extract skeleton features, and compare performances; (iv) we provide ablation studies to show the contribution of different components of our approach, including the RMP module. 
% We surveyed and compared more point-based algorithms. 2) We analysed the utilization of each human key point and explore the impact of different input key points on the results. 3) We achieved better performance by adopting RMP module and improving the loss function. 

\section{Related Work}\label{sec:related}
We will first briefly review appearance-based and model-based gait recognition approaches. We will then summarize 3D point cloud processing methods, which inspired this work.
%Most of existing gait recognition methods learn representations from analysis of the skeletons or silhouettes of subjects, therefore, gait recognition approaches can be classified into two categories: appearance-based and model-based. In this section, we will briefly review these two sorts of methods and we also introduce point cloud processing algorithms that inspired this work .
%
%
%
\subsection{Gait Recognition}
\subsubsection{Appearance-based Gait Recognition}
Appearance-based methods~\cite{fan2020gaitpart,chao2019gaitset,lin2020learning,sepas2020view,zhang2019cross,sepas2021gait,hou2020gait} rely on silhouettes to represent human body. Silhouettes can be extracted by using background subtraction followed by binarization or using the chroma key compositing. The simplest recognition approach is to directly measure the similarity (e.g., Euclidean or cosine distance) between gallery sequences and a probe sequence~\cite{sarkar2005humanid}, which is called baseline algorithm in gait recognition community. Another appearance-based approach~\cite{han2005individual} obtains a Gait Energy Image (GEI) by aggregating gait information over a complete gait cycle into a single image, and using machine learning techniques~\cite{wu2016comprehensive,yu2017invariant} to extract gait features. There are other works that directly employ a sequence of silhouettes as input, instead of using their average, and employ 3D Convolutional Neural Network  (CNN)-based~\cite{wolf2016multi} or Long-short Term Memory (LSTM)-based~\cite{tong2018multi} methods to extract features. These approaches can preserve more temporal and spatial information but incur higher computational costs and are harder to train. Other approaches~\cite{fan2020gaitpart,chao2019gaitset,lin2021gait} treat silhouettes as an unordered set %consisting of independent silhouettes rather than a continuous set of silhouettes. 
These approaches do not require temporal information, and have less computational cost.

\subsubsection{Model-based Gait Recognition} 
Model-based methods extract gait information by modeling human body structure or motion patterns of different body parts, and they mainly take skeletons as input. Skeletons can be obtained by high accuracy depth sensors or pose estimation algorithms. If skeleton data is highly accurate, model-based methods could be more robust to appearance changes and camera viewpoints, but it is a challenging task with increased computational complexity.
%, making Thus model-based methods are not as popular as appearance-based methods. 

An early model-based work~\cite{tanawongsuwan2001gait} marks different human body parts manually or by specific sensors to obtain human joint positions. Pose-based Temporal-Spatial Network (PTSN)~\cite{liao2017pose} uses pose estimation to extract body joint data with a two-branch network, which extracts temporal features via LSTM and spatial features with a CNN. PoseGait~\cite{liao2020model} uses a 3D pose estimation method to get 3D key-points, and employs handcrafted features (joint angle, limb length and joint motion) based on the euclidean distance between these joints. It then obtains spatio-temporal features with CNN. Recently, researchers proposed to adopt graph convolutional networks (GCN) to extract gait features from skeletons~\cite{li2020jointsgait,teepe2021gaitgraph}. They create gait graph structure from raw silhouettes and then extract human motion features by GCN.

More recently, MSGG~\cite{peng2021learning} was proposed to extract motion features from skeleton data, and then fuse motion features with contour features obtained by GaitPart~\cite{fan2020gaitpart}. Doing so, MSGG achieved better performance than the original GaitPart. Yet, MSGG is developed based on ST-GCN~\cite{yan2018spatial} and ResGCN~\cite{song2020stronger} and its training is computationally expensive. The parts for learning human contour features and skeleton key point features are pre-trained separately and then combined for final fine-tuning. \cite{wang2022two} is another GCN-based method, which also employs GaitPart for silhouette representation. However, the complexity of the key point processing methods in \cite{peng2021learning} and \cite{wang2022two} may limit their generalizability and widespread use.
%Some recent approaches~\cite{peng2021learning,wang2022two} 
%%fuse appearance-based and model-based, 
%combine silhouette and skeleton features and obtain better performance, which show that these two kinds of data are complementary, and provide more comprehensive representation of gait.
%and their combination is more distinguish and comprehensive for gait recognition. 

\subsection{Point Cloud Analysis}
Point cloud data is a set of 3D points obtained by sensors, such as a depth camera and Lidar. Different from 2D images, 3D point cloud data is unordered and irregular. Thus, traditional CNN-based methods cannot be readily applied to point clouds. To address this issue, PointNet~\cite{qi2017pointnet} employs a max-pooling layer and shared multi-layer perceptrons (MLP) to obtain permutation invariant features. 
%for the downstream tasks, e.g., classification and segmentation. 
Later PointNet++~\cite{qi2017pointnet++} was introduced, which adopts 
%However, PointNet does not capture the local information between points. Hence later, the same group proposed PointNet++~\cite{qi2017pointnet++}, which is developed based on PointNet. It uses 
Farthest Point Sampling (FPS) to group a set of key points as local neighbors in each layer. DGCNN~\cite{wang2019dynamic} uses a dynamic graph CNN to aggregate each point's non-local neighbor features by linking each center point with its $k$ nearest neighbors in each layer. There are other works~\cite{maturana2015voxnet,le2018pointgrid}, which, instead of using 3D point clouds directly as input, first transform an unstructured 3D point cloud data into voxel grids, and then use 3D convolution to extract features for classification or segmentation tasks.
\section{Motivation}\label{sec:motivation}
Before introducing the details of our approach, we first provide the motivation behind the GaitPoint+ in this section.

\noindent{\textbf{Why combine skeleton and silhouette features?}}
In vision-based gait recognition, an individual's gait video $V$ (the input), processed by computer vision algorithms, is matched to an existing/known individual's gait video $V^{'}$ in a database to recognize or identify the person. However, the appearance of the unknown individual might be different from the videos in the database for various reasons, such as
%or exterior variations might be changed, e.g., 
different clothing, carrying some items or changing camera views. These changes present additional challenges to existing appearance-based gait recognition approaches. Although CNNs have been shown to be effective in extracting features from images and videos, they are prone to overfitting to the contour or shape of an individual rather than capturing the walking motion. Therefore, appearance changes can significantly affect the performance of CNN- or appearance-based methods. 

\begin{figure}[h]
\centering
\vspace{-0.2cm}
\includegraphics[width=0.45\textwidth]{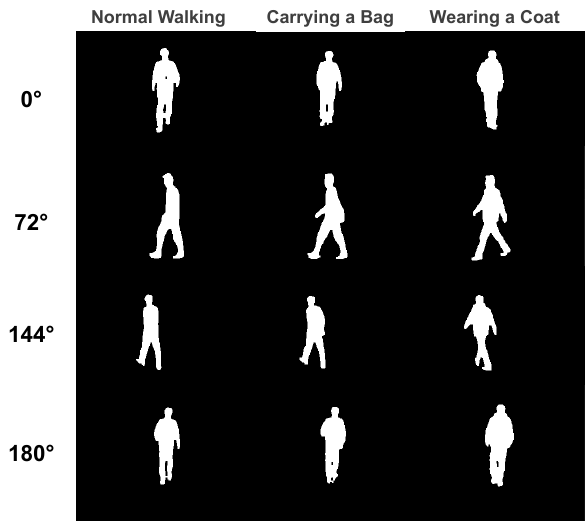}
\vspace{-0.4cm}
\caption{\small{An individual's silhouette from different angles when walking normal, carrying a bag and wearing a coat.}}
\label{fig:walking}
\vspace{-0.4cm}
\end{figure}

Table~\ref{tab: performance} shows performances of several SOTA appearance-based methods for three different appearances, namely normal walking, walking while carrying a bag and walking while wearing a coat, on CASIA-B dataset~\cite{yu2006framework}. These results were obtained by running the codes published by respective authors on our local machine with only one GPU and a batch size of $4 \times 8 = 32$, which covers 8 videos each from 4 people. In the original papers, results are obtained with a batch size of $8 \times 16 = 128$. 
%where 8 is the number of picked people and 16 is the number of videos each person has in the batch. However, the batch size of all our experiments is $4 \times 8$, eight videos are picked from each of four different people. 
Thus, the results presented here can be slightly different from the ones reported in the original papers. As shown in Table \ref{tab: performance}, for all the methods, the highest accuracy is obtained for `Normal walking'. The performance drops significantly if a person carries a bag, and the performance is the worst when a person wears a coat. This shows that appearance changes can have a significant negative impact on recognition accuracy. To further analyze the reasons behind this phenomenon, we visualize the silhouettes of the same person for three walking statuses from different view angles in Fig.~\ref{fig:walking}. As can be seen, when carrying a backpack, the contour is similar to the contour for `Normal Waking', especially from front and and back views (0$^{\circ}$ and 180$^{\circ}$). When wearing a coat, on the other hand, the upper body looks larger for all view angles. This illustrates why all the models in Table.~\ref{tab: performance} have better performance for carrying a bag than wearing a coat. Thus, purely appearance- or CNN-based methods are, in general, highly dependent on appearance or silhouette information, causing them not to learn walking information well.
\begin{table}[t!]
\centering
\caption{\small{Performance of SOTA appearance-based gait recognition methods for different walking scenarios with varying appearances}}
\vspace{-0.2cm}
\resizebox{1\linewidth}{!}{
\begin{tabular}{|c|c|c|c|}
\hline
         & Normal Walking & Carrying Bag & Wearing Coat \\ \hline
GaitSet~\cite{chao2019gaitset}  & 95.46\% & 89.02\%       & 73.82\%        \\ \hline
GaitPart~\cite{fan2020gaitpart} & 95.5\% & 90.6\%       & 76.55\%        \\ \hline
GaitGL~\cite{lin2021gait}   & 96.21\% & 92.72\%       & 80.93\%        \\ \hline
\end{tabular}
}
\vspace{-0.2cm}
\label{tab: performance}
\end{table}

To remedy this issue, skeleton- or model-based approaches~\cite{teepe2021gaitgraph, gao2022gait, zhang2022spatial} have been proposed. 
%Yet, since human shape or contour information is mostly discarded during pose key point detection, the performance of most skeleton-based methods are lower than most silhouette- and CNN-based approaches. 
GaitGraph~\cite{teepe2021gaitgraph} is a skeleton-based method, which uses HRNet~\cite{sun2019deep} to obtain the human pose key points. Then, it employs ResGCN~\cite{song2020stronger} to extract human motion features based on those key points. However, since most human contour information is filtered out during pose key point detection, the performance of GaitGraph was lower than most image- and convolution-based methods~\cite{chao2019gaitset,fan2020gaitpart,lin2021gait,hou2020gait}. For heavy clothing scenarios, skeleton-based methods perform relatively better than silhouette-based ones, showing that these two data modalities are complementary and provide more distinguishing features together for different walking conditions. As discussed in Sec.~\ref{sec:related}, MSGG~\cite{peng2021learning} fuses motion and contour features, and provides better performance than the  GaitPart showing once more that the combination of the two data modalities is indeed helpful for gait recognition. Yet, as mentioned before,
%MSGG is developed based on ST-GCN~\cite{yan2018spatial} and ResGCN~\cite{song2020stronger} and its training is computationally expensive. The parts for learning human contour features and skeleton key point features are pre-trained separately and then combined for final fine-tuning. \cite{wang2022two} is another GCN-based method, which also employs GaitPart for silhouette representation. However, 
the complexity of the key point processing methods in \cite{peng2021learning} and \cite{wang2022two} may limit their generalizability.

To address all the aforementioned issues and motivated by the above discussion, our goals are (i) to design a one-stage, end-to-end gait recognition method, (ii) to utilize human pose key points as auxiliary information by combining them with silhouette features, and (iii) to  perform this without incurring significant additional computation cost and without requiring pre-training of each branch. More specifically, the key point learning module should be lightweight. Since appearance or CNN-based methods already incur considerable amount of computation, it is preferable that the key point learning module is faster and has lighter weight.

\section{Proposed Method}
To utilize other features, which are more robust to appearance changes than silhouette features, and address the issues of purely silhouette- and CNN-based methods, we propose a novel, one-stage, end-to-end framework for gait recognition, referred to as the GaitPoint+. Our approach aggregates silhouette and skeleton features, and boosts the performance of purely appearance-based methods without incurring significant additional computational cost. In other words, our 3D point processing approach provides the performance improvement without requiring the use of heavier network models and strikes a desirable balance between accuracy and efficiency. 

\subsection{GaitPoint+ with Recycling Max-Pooling}\label{ssec:GaitPointplusRMP}
\subsubsection{GaitPoint+ Framework} 
\begin{figure*}[!th]
\centering
\includegraphics[width=0.9\textwidth]{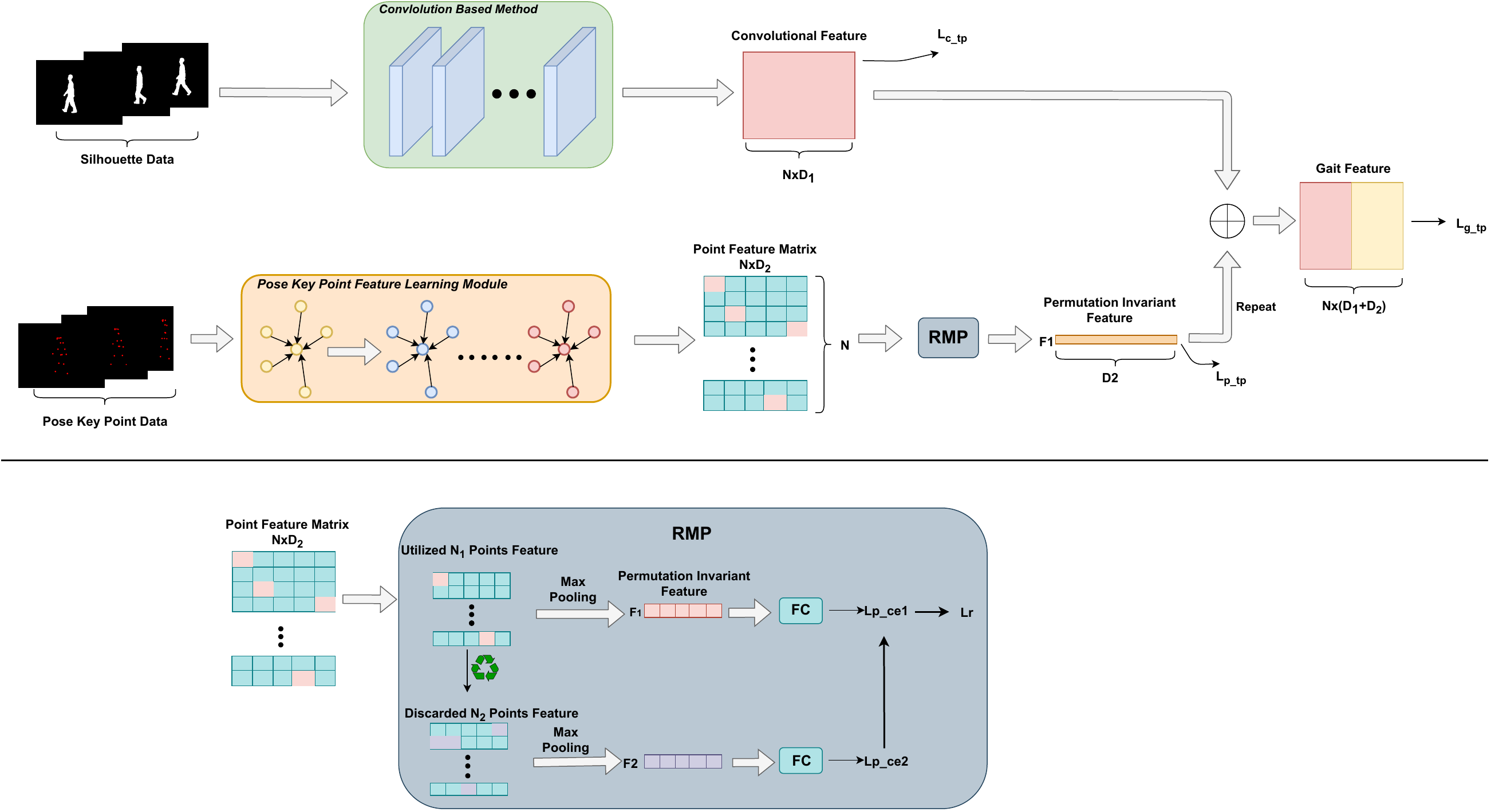}
\vspace{-0.25cm}
\caption{\small{The network structure of the GaitPoint+ with Recycling Max-pooling (RMP) Module. $L_{p}\__{tp}$, $L_{c}\__{tp}$ and $L_{g}\__{tp}$ are the triplet losses for the permutation invariant, convolutional and gait features, respectively. $F_1$ and $F_2$ are the permutation invariant features obtained by max-pooling from original key points and discarded key points, respectively. $L_{p}\__{ce1}$ and $L_{p}\__{ce2}$ are the cross entropy losses for the permutation invariant $F_1$ and $F_2$, respectively. $L_r$ is refinement loss. FC represents Fully Connected Layer, and two FC layers share parameters.}}
\vspace{-0.3cm}
\label{fig:model}
\end{figure*}

%To utilize other features, which are more robust to appearance changes than silhouette features, and address the issues of purely silhouette- and CNN-based methods, we propose a novel and end-to-end framework for gait recognition, referred to as the GaitPoint+. 
GaitPoint+ incorporates both skeleton and silhouette features, and employs a Recycling Max-Pooling (RMP)~\cite{chen2022discard} module to recycle the human key points, which would have been discarded otherwise due to traditional max-pooling while processing skeletons. The overall architecture of GaitPoint+ is presented in Fig.~\ref{fig:model}. For the upper branch, the input is a sequence of silhouettes, which can be processed by any silhouette- or convolution-based approach. In our work, without loss of generality, we employ and compare GaitSet~\cite{chao2019gaitset}, GaitPart~\cite{fan2020gaitpart} and GaitGL~\cite{lin2021gait} in this upper branch. The model performs pooling operation on feature maps multiple times, to obtain features in different receptive fields, and outputs the convolutional features $C \in \mathbf{R}^{N\times D_2}$, where $N$ is the number of output features, and $D_2$ is the dimension of each feature. 

For the lower branch, the input is a sequence of 2D pose key point data, which are obtained by HRNet~\cite{sun2019deep} from each frame. We first convert the key point data into a 3D point cloud as follows: \vspace{-0.1cm}
\begin{equation} \label{eq:PointForm}
    \{(x_{it},y_{it},t)\,\,|\,\,i\in \{1,...,K\}, t\in\{0,..,T-1\}\},
\end{equation}
where $K$ refers to the number of key points in each frame and $T$ is the number of frames. Then, the Pose Key Point Feature Learning Module extracts the distinguishing features from the skeleton key point cloud. Although this module can be any point cloud analysis method in general, we employ PointNet in this work for the reasons explained in detail below.
\begin{table*}[tb!]
\centering
\caption{Comparison of ResGCN and PointNet in terms of size, running speed and accuracy. SD is the standard deviation of the accuracy for different walking scenarios.}
\resizebox{0.72\textwidth}{!}{
\begin{tabular}{|c|ccc|cccc|}
\hline
\textbf{Model} &
\multicolumn{3}{|c|}{\textbf{Model Attribute}}                                                                                                                       & \multicolumn{4}{c|}{\textbf{Model   Accuracy}}                                                                                          \\ \hline
\multicolumn{1}{|c|}{}         & \multicolumn{1}{c|}{Forw./backw.} & \multicolumn{1}{c|}{Estim. Total} & Running Time/ & \multicolumn{1}{c|}{Normal}   & \multicolumn{1}{c|}{Carrying}    & \multicolumn{1}{c|}{Wearing}   & SD              \\ 
\multicolumn{1}{|c|}{}         & \multicolumn{1}{c|}{pass size (MB)} & \multicolumn{1}{c|}{Size(MB)} & Batch (ms) & \multicolumn{1}{c|}{Walking}   & \multicolumn{1}{c|}{Bag}    & \multicolumn{1}{c|}{Coat}   &               \\ \hline
\multicolumn{1}{|c|}{PointNet~\cite{qi2017pointnet}} & \multicolumn{1}{c|}{\textbf{2541.75}}                  & \multicolumn{1}{c|}{\textbf{2546.52}}           & \textbf{19.07}           & \multicolumn{1}{c|}{\textbf{64.28\%}} & \multicolumn{1}{c|}{\textbf{56.18\%}} & \multicolumn{1}{c|}{\textbf{51.92\%}} & \textbf{5.13\%} \\ \hline
\multicolumn{1}{|c|}{ResGCN~\cite{song2020stronger}}   & \multicolumn{1}{c|}{3604.76}                           & \multicolumn{1}{c|}{3607.71}                    & 34.11                    & \multicolumn{1}{c|}{60.85\%}          & \multicolumn{1}{c|}{51.07\%}          & \multicolumn{1}{c|}{40.95\%}          & 8.12\%          \\ \hline
\end{tabular}}
\vspace{-0.2cm}
\label{tab:comp}
\end{table*}
\\ \textbf{Why employ PointNet?}
%\label{ssec:whypoint}
As mentioned above, while both methods in \cite{peng2021learning, wang2022two} combine silhouette and skeleton features, they are developed based on GCN, and their training is computationally expensive. Moreover, the complexity of their key point processing methods may limit their generalizability and widespread use. Since our proposed model is one-stage, and our goal is to design an efficient model, its auxiliary human key point processing module needs to satisfy the following: (i) the skeleton key point learning module should be compatible with the learning strategy of most silhouette- and convolution-based methods. For example, GaitGraph and MSGG are based on ResGCN, which is trained with contrastive learning~\cite{chen2020simple}. It extracts three sequences (joint, velocity and bone) from the original skeleton sequence, and feeds them to the model, followed by fine-tuning. This training strategy is very different from the ones used by silhouette-based methods, which minimize the walking feature distance for the same person while maximizing it for different people; (ii) the key point feature extraction process should be lightweight and faster. 

To analyze which model is a better fit as the human key point processing module, we first compare ResGCN~\cite{song2020stronger} with PointNet~\cite{qi2017pointnet} in terms of model size, speed, and accuracy. Both networks are trained and evaluated in the same way as the silhouette-based methods~\cite{chao2019gaitset, fan2020gaitpart, lin2021gait} on the CASIA-B dataset~\cite{yu2006framework}. The results, summarized in Table~\ref{tab:comp}, show that PointNet has smaller memory requirement, is faster, and provides better accuracy at the same time compared to ResGCN. Moreover, PointNet is more robust to appearance variations having a lower standard deviation in accuracy for three different walking scenarios. Thus, in our method, we employ PointNet for pose key point learning. Additional ablation study results are provided in Sec.~\ref{ssec:Abla_pointnet} comparing different point-based methods.

As shown in Fig.~\ref{fig:model}, permutation invariant features $F \in \mathbf{R}^{D_1}$, obtained from PointNet in the lower branch, contain  motion/walking information that is more robust to appearance changes. The vector $F$ is replicated $N$ times, to obtain an $N \times D_{1}$ vector, which is then concatenated with convolutional features $C$ to obtain the overall gait features $G \in \mathbf{R}^{N \times (D_{1}+D_{2})}$.

\subsubsection{Recycling Max-Pooling Module and Point Utilization Analysis}
\begin{figure}[hb!]
\centering
\vspace{-0.5cm}
\includegraphics[width=0.9\linewidth, height=0.6\linewidth]{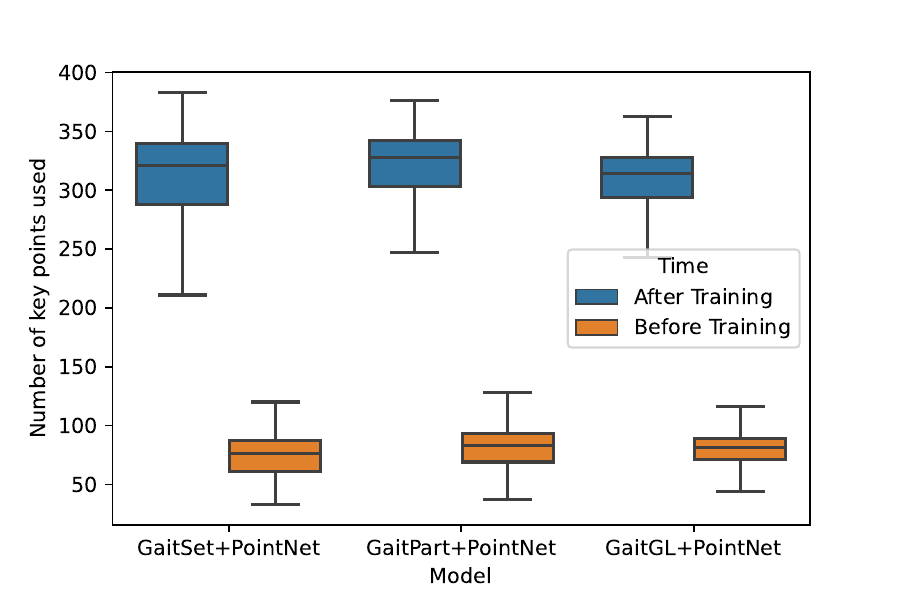}
\vspace{-0.4cm}
\caption{\small{Box plots of the number of human key points used when PointNet is combined with three silhouette-based models. The total number of input key points is 1020 (60 frames and 17 key points in each frame). There are about 75 points used by all models before training. This number increases to 325 when networks are well trained.}}
\label{fig:total_point_used}
\vspace{-0.1cm}
\end{figure}

%\subsubsection{\textbf{Why use RMP module?}}\label{ssec:whyrmp}
%Point cloud is a set of unstructured and unordered data, since 
PointNet and most existing point-based point cloud processing models apply the max-pooling operation to get permutation invariant features, and only use features from a portion of points while discarding others. Chen et al.~\cite{chen2022discard} showed that these discarded points still contain useful information and proposed an RMP module to recycle them. Inspired by this, we first investigate the number of skeleton key points utilized when SOTA silhouette-based methods, namely GaitSet, GaitPart and GaitGL, are used together with PointNet on the CASIA-B dataset. The results presented in Fig.~\ref{fig:total_point_used} show that, in all experiments, the number of utilized points increases at the end of training compared to before training. This indicates that the network is learning to pick up more points to better represent a person's gait. Thus, we apply the RMP module to increase the point utilization further by recycling the discarded points, and output a set of point features that can better describe an individual's walking pattern. 

We also analyze the utilization of every human key point after one max-pooling in Sec.~\ref{sec:Ablation} and show that traditional max-pooling discards a lot of the hip points, e.g., and these points may still contain useful information, which we aim to utilize via recycling through an RMP module.

% Motivated by the findings in Sec.~\ref{ssec:whyrmp}, we deploy RMP module~\cite{chen2022discard} in the point cloud analysis model to increase the utilization of human key points by recycling the discarded points.
The RMP module is shown at the bottom of Fig.~\ref{fig:model}. $F_1 \in \mathbf{R}^{D_1}$ is obtained from the traditional max-pooling operation, which selects the most representative points from the raw key points, and discards the rest of the points. Hence, we perform a second max-pooling operation among the discarded key points to get another permutation invariant feature set $F_2$ for training. To combine $F_1$ and $F_2$, we use a hierarchical loss to refine $F_1$. This loss function is incorporated with classification loss and refinement loss. The classification losses $L_{p}\__{ce1}$ and $L_{p}\__{ce2}$ are cross-entropy losses and obtained by adding fully connected layers after $F_1$ and $F_2$, respectively. $L_{p}\__{ce1}$ and $L_{p}\__{ce2}$ are calculated based on $\hat{y_i}$ and $y_i$, where $y_i$ is the one-hot encoded ground truth, and $\hat{y_i}$ is the soft-max prediction obtained based on $F_1$ and $F_2$. 

Refinement loss is designed to use feature set $F_2$ to decrease the classification loss of $F_1$, and is defined as follows:
\begin{equation}
    L_r=| 1-e^{(L_{p}\__{ce2}-\rho L_{p}\__{ce1})}  |
    \label{eq:refinement-loss}
\end{equation}
\begin{equation}
    \rho=\alpha\cdot e^{(\sum_{m=1}^k y_m \cdot \hat{y_m})},
    \label{eq:rhodefinition}
\end{equation}

\noindent where $\alpha>1$, $k$ is the number of classes, and $y_m$ and $\hat{y_m}$ are the ground truth and the prediction based on $F_1$, respectively. Chen et al.~\cite{chen2022discard} showed that the performance of permutation invariant features kept after first max-pooling ($F_1$) is slightly better than that of $F_2$. Thus, it can be inferred that $L_{p}\__{ce2}>L_{p}\__{ce1}$.  $L_r$ in Eq.~(\ref{eq:refinement-loss}) is minimized when $L_{p}\__{ce2} \approx \rho_i L_{p}\__{ce1}$, where $\rho$ is defined as in Eq.~(\ref{eq:rhodefinition}). At the beginning of the training, the neural network has not yet learned useful information, which might cause the prediction to be inaccurate, resulting in $\sum_{m=1}^k y_m \cdot \hat{y_m} \approx 0$, and $\rho_i \approx \alpha_i$. At this stage, since $\rho_i$ is small, network pays more attention to the classification based on $F_1$ and $F_2$ respectively, and refinement loss does not have much effect. As training continues, the value of $\rho$ is increased, $L_{p}\__{ce1}$ is pushed to be smaller by $L_{p}\__{ce2}$, and $F_1$ starts to be refined by $F_2$.

% However, too large $L_p\__c_e_2$ is not desirable, an absolute value is applied in Eqn.~\ref{eq:refinement-loss} to restrict $L_p\__c_e_2$ and $L_r>=0$. 

% \begin{figure*}[!th]
% \centering
% \includegraphics[width=0.9\textwidth]{image/RMP.pdf}
% \vspace{-0.3cm}
% \caption{\small{The structure of the RMP Module. $F_1$ and $F_2$ are the permutation invariant feature obtained by max-pooling operation from original human key points and discarded key points. $L_p\__c_e_1$ and $L_p\__c_e_2$ are the cross entropy losses for the permutation invariant $F_1$ and $F_2$, respectively. $L_r$ is refinement loss. FC represents Fully Connected Layer, and the two FC layers share parameters }}
% \vspace{-0.3cm}
% \label{fig:model}
% \end{figure*}

\subsubsection{Loss Function}
%Triplet loss, cross-entropy loss and refinement loss are employed when training GaitPoint+. 
%\hl{The loss function used in the RMP module are the cross-entropy and refinement loss. The RMP loss is computed as}
The loss function for the RMP module ($L_{rmp}$) is computed as
\begin{equation}
{
    L_{rmp}=L_{p}\__{ce1}+L_{p}\__{ce2}+L_r},
\end{equation}
where $L_{p}\__{ce1}$ and $L_{p}\__{ce2}$ are the cross-entropy losses for the original permutation-invariant and the recycled permutation-invariant features, respectively. $L_r$ denotes the refinement loss.
The total loss is formed by triplet loss and RMP loss:
\begin{equation}
{
    L=L_{c}\__{tp}+L_{p}\__{tp}+L_{g}\__{tp}+L_{rmp}},
\label{eq:final_loss}
\end{equation}
where $L_{p}\__{tp}$, $L_{c}\__{tp}$ and $L_{p}\__{tp}$ are the triplet losses for the permutation invariant, convolutional and gait features, respectively. The triplet loss is $L_{tp}=max(D(A,P)-D(A,N)+M,0)$, where $D(A,P)$ and $D(A,N)$ are the distances of the anchor to a positive sample and a negative sample, respectively. $M$ is the margin value set as 0.2, which is used to control the distance between positive and negative samples. With the loss $L$, the distance between the features from the videos of the same person is decreased, while the distance between the features from the videos of different people is increased.
\section{Experimental Results}\label{sec:exp}
\subsection{Dataset and Training Details}
\subsubsection{Dataset}
CASIA-B~\cite{yu2006framework} is a commonly-used and publicly available gait recognition dataset.
%created by the Institute of Automation, Chinese Academy of Sciences (CASIA). 
It contains videos from 124 subjects. For each subject, each video is captured from 11 different views, which
%Video of  each subject contains 11 views captured from 11 cameras located at 11 different views at the same time. 
are evenly distributed between $0^{\circ}$ and $180^{\circ}$, with $18^{\circ}$ increments. Each view contains 10 sequences, where six videos are normal walking sequences (NM), two are  walking with a bag sequences (BG) and remaining two are walking in a coat sequences (CL). In total, each subject has $11\times (6+2+2) = 110$ sequences. We use HRNet~\cite{sun2019deep} to obtain 17 human key points as the input of skeleton processing branch.
%which are nose, left eye, right eye, left ear, right ear, left shoulder, right shoulder, left elbow, right elbow, left wrist, right wrist, left hip, right hip, left knee, right knee, left ankle, right ankle. 
In the experiments, we follow the most popular training protocol, namely large-sample training~\cite{chao2019gaitset}, where the first 74 subjects and the remaining 50 subjects are used for training and testing, respectively. In the testing set, 4 out of the 6 NM sequences are placed into the gallery set, while the remaining 2 NM sequences and the sequences of BG and CL are put into the query set.

\subsubsection{Training Details}
\label{sssec: settings}
For the upper branch of our method, we use different SOTA silhouette-based approaches, namely GaitSet, Gaitpart and GaitGL, and compare the performance with or without our point-based approach and RMP module. The input silhouettes are resized and aligned to resolution of $64\times44$. For the lower branch, the input is a sequence of raw skeleton data with the size of $60\times17\times3$, since there are 60 frames for each sequence, with 17 human key points for each frame and each key point having 3 channels (horizontal and vertical coordinates and time stamp). The batch size is set as 32 and the maximum number of iterations is set to 80000. At each training iteration, 4 people (out of 74) are randomly chosen, and 8 videos are randomly picked from each subject’s video pool, resulting in a total of $4\times8=32$ videos for each step. The learning rate is set to $1e-4$ and the momentum is set to 0.9. We set $K$, $M$ and $T$ as 17, 0.2 and 60, respectively. All experiments are performed on one GPU with exactly the same training and testing environment, e.g., the same optimizer, batch size, learning rate etc. 
\subsection{Results}
\begin{table*}[ht!]
\centering
\caption{\small{Averaged rank-1 accuracy on CASIA-B dataset under three different walking scenarios, excluding identical-view cases. N is the number of viewing angles, for which the accuracy is improved. \textbf{Bold} and \underline{Underline} indicate the best and 2nd best performances.}}
\resizebox{\textwidth}{!}{
\begin{tabular}{|c|c|c|c|c|c|c|c|c|c|c|c|c|c|c|}
\hline
                    &                   & 0°               & 18°              & 36°              & 54°              & 72°              & 90°              & 108°             & 126°             & 144°             & 162°             & 180° & N             & Mean    \\ \hline
\multirow{6}{*}{NM} & GaitSet\cite{chao2019gaitset}           & 91.80\%          & \textbf{97.70\%}          & \textbf{98.60\%}          & \textbf{97.50\%}          & 93.30\%          & 92.20\%          & 94.20\%          & \underline{96.90\%}         & \textbf{98.40\%}          & \underline{97.00\%}          & 87.60\%   &/       & 95.02\% \\ 
                     & GaitSet+PN(ours)  & \underline{94.24\%} & \underline{97.07\%}          & \underline{96.60\%}         & \underline{95.96\%} & \underline{96.43\%} & \underline{95.00\%} & \underline{95.67\%} & 95.10\%          & 97.40\%         & 96.90\% & \underline{91.30\%} &5 & \underline{95.61\%} (↑0.59\%) \\
                    & GaitSet+PN+RMP(ours)  & \textbf{94.75\%} & 96.47\%          & 96.10\%          & 95.86\% & \textbf{96.94\%} & \textbf{96.96\%} & \textbf{96.70\%} & \textbf{97.40\%}          & \underline{97.70\%}         & \textbf{97.60\%} & \textbf{93.30\%} &7 & \textbf{96.34\%} (↑1.32\%) \\
                    \cline{2-15} 
                    & Gaitpart\cite{fan2020gaitpart}          & 92.10\%          & \textbf{97.10\%}          & \textbf{98.60\%}          & \textbf{96.50\%}          & 93.60\%          & 92.20\%          & 94.70\%          & \textbf{97.70\%}          & \textbf{97.30\%}         &\textbf{97.30\%}          & 88.60\% &/         & \textbf{95.06\%} \\ 
                     & GaitPart+PN  & \underline{93.84\%} & 94.14\%          & \underline{95.40\%}          & \underline{94.95\%} & \underline{95.51\%} & \underline{93.80\%} & \textbf{96.19\%} & \underline{96.20\%}          & 95.50\%         & 95.20\% & \underline{87.30\%} &4 & 94.37\% (↓0.69\%) \\
                    & GaitPart+PN+RMP(ours) & \textbf{94.24\%} & \underline{95.66\% }        & 94.90\%          & 93.33\%          & \textbf{95.72\%} & \textbf{96.20\%} & \underline{95.98\%} & 95.70\%          & \underline{96.80\%}          & \underline{96.70\%} & \textbf{89.80\%} &5 & \underline{95.00\%} (↓0.06\%) \\ \cline{2-15} 
                    & GaitGL\cite{lin2021gait}            & 94.20\%          & \underline{96.60\%}          & \textbf{98.40\%}          & \textbf{97.00\%}          & \textbf{96.00\%}          & 92.30\%          & 95.20\%          & \textbf{98.20\%}          & \textbf{98.00\%}          & \underline{96.50\%}          & \underline{92.50\%}         &/ & \textbf{95.90\%} \\ 
                    & GaitGL+PN(ours)   & \textbf{96.16\%} & 94.01\%         & \underline{95.30\%}          & 94.45\%          & \underline{95.61\%} & \textbf{97.06\%} & \underline{96.39\%} & \underline{97.30\%}          & \underline{97.00\% }         & \textbf{96.80\%}          & \textbf{92.60\%} &5 & 95.70\% (↓0.20\%) \\
                    & GaitGL+PN+RMP(ours)   & \underline{94.65\%} & \textbf{96.77\%}          & 94.80\%          & \underline{94.55\%}          & 95.31\% & \underline{96.42\%} & \textbf{96.91\%} & 96.10\%          & 96.60\%          & 96.00\%          & 91.70\% &4 & \underline{95.88\%} (↓0.02\%) \\ \hline \hline
\multirow{6}{*}{BG} & GaitSet\cite{chao2019gaitset}           & 83.60\%          & 90.00\%          & \textbf{94.55\%}          & 92.65\%          & 86.30\%          & 82.60\%          & 85.80\%          & 90.40\%          & 91.60\%          & 91.42\%          & 80.00\%         &/   & 88.08\% \\ 
                    & GaitSet+PN(ours)  & \textbf{92.02\%} & \textbf{93.06\%}          & 91.94\%          & \textbf{93.54\%} & \underline{89.90\%} & \underline{88.49\%} & \underline{89.08\%} & \textbf{93.40\%}          & \textbf{95.70\%}         & \underline{94.45\%} & \underline{85.70\%} &10 & \textbf{91.57\%} (↑3.49\%) \\
                    & GaitSet+PN+RMP(ours)  & \underline{90.41\%} & \underline{91.33\%} & \underline{91.94\%}          & \underline{93.44\%} & \textbf{90.93\%} & \textbf{89.46\%} & \textbf{90.72\%} & \underline{92.70\%} & \underline{95.50\%} & \textbf{94.45\%} & \textbf{86.10\%} &10 &\underline{91.54\%} (↑3.46\%) \\ \cline{2-15} 
                    & Gaitpart\cite{fan2020gaitpart}          & 86.50\%          & \textbf{92.83\%}          & \textbf{94.85\%}          & \textbf{92.55\%}          & 89.10\%          & 83.50\%          & 88.60\%          & \underline{92.40\%}          & \textbf{93.90\%}          & \underline{91.11\%}          & 81.20\%         &/ & \underline{89.69\%} \\  
                    & GaitPart+PN(ours)  & \underline{88.08\%} & 89.80\%          & \underline{92.45\%}          & 91.35\% & \underline{90.21\%} & \underline{86.77\%} & \textbf{91.43\%} & 91.50\%          &92.90\%         & 89.80\% & \underline{81.20\%} &4 & 89.59\% (↓0.10\%) \\
                    & GaitPart+PN+RMP(ours) & \textbf{89.70\%} & \underline{91.63\%}          & 91.84\%          & \underline{92.19\%}          & \textbf{91.13\%} & \textbf{89.89\%} & \underline{90.41\%} & \textbf{92.70\%}          & \underline{93.70\%} & \textbf{93.74\%} & \textbf{83.50\%} &7 & \textbf{90.95\%} (↑1.26\%) \\ \cline{2-15} 
                    & GaitGL\cite{lin2021gait}            & 91.00\%          & \textbf{94.75\%}          & \textbf{96.36\%}          & \textbf{94.39\%}          & 92.00\%          & 85.30\%          & 86.70\%          & \textbf{94.70\%}          & \textbf{96.30\%}          & 95.56\%          & 85.80\%         &/ & 92.08\% \\ 
                    & GaitGL+PN(ours)   & \underline{93.43\%} & 92.76\%          & 92.55\%          & \underline{94.06\%}          & \underline{93.61\%} & \underline{92.90\%} & \underline{93.47\%} & \underline{94.60\%}          & \underline{96.20\%}          & \textbf{96.26\%}          &\textbf{88.50\%} &6 & \underline{93.49\%} (↑1.41\%) \\
                    & GaitGL+PN+RMP(ours)   & \textbf{93.94\%} & \underline{93.98\%}          & \underline{93.47\%}          & 91.77\%          & \textbf{94.02\%} & \textbf{93.66\%} & \textbf{93.88\%} & 93.90\% & 94.30\%          & \underline{95.96\%} & \underline{87.80\%} &6 & \textbf{93.59\%} (↑1.51\%) \\ \hline \hline
\multirow{6}{*}{CL} & GaitSet\cite{chao2019gaitset}           & 65.30\%          & 76.60\%          & 79.70\%          & 76.30\%          & 74.20\%          & 69.60\%          & 70.60\%          & 75.10\%          & 76.30\%          & 73.90\%          & 56.00\%       &/    & 72.15\% \\ 
                    & GaitSet+PN(ours)    & \underline{79.30\%}          &\underline{86.10\%}          & \underline{85.10\% }         & \underline{84.39\%}          & \underline{82.08\%}          & \underline{80.11\%}          & \underline{82.92\%}          & \underline{84.70\%}          & \underline{84.90\%}          & \underline{84.00\%}          & \underline{76.00\%}       &11    & \underline{82.69}(↑10.54\%) \\
                    & GaitSet+PN+RMP(ours)  & \textbf{84.30\%} & \textbf{86.80\%} & \textbf{84.90}\% & \textbf{84.59\%} & \textbf{83.33\%} & \textbf{82.05\%} & \textbf{84.79\%} & \textbf{86.10\%} & \textbf{86.80\%} & \textbf{84.90\%} & \textbf{78.40\%} &11 & \textbf{84.27\%} (↑12.12\%) \\ \cline{2-15} 
                    & Gaitpart\cite{fan2020gaitpart}          & 67.40\%          & 81.50\%          & 83.60\%          & 78.50\%          & 72.90\%          & 68.90\%          & 73.40\%          & 80.70\%          & 79.10\%          & 75.40\%          & 62.00\%         &/ & 74.86\% \\ 
                    & GaitPart+PN(ours)    & \underline{80.60\%}          &\underline{83.20\%}          & \underline{84.70\% }         &\textbf{86.43\%}           & \underline{83.12\%}   & \underline{78.98\%}          & \underline{83.96\%}          & \underline{85.10\%}          & \underline{81.70\%}          & \underline{82.50\%}    & \underline{72.00\%}   &11    & \underline{82.03\%}(↑7.17\%) \\
                    & GaitPart+PN+RMP(ours) & \textbf{85.10\%} & \textbf{85.50\%} & \textbf{85.60\%} & \underline{85.61\%} & \textbf{84.58\%} & \textbf{82.28\%} & \textbf{84.69\%} & \textbf{87.50\%} & \textbf{87.90\%} & \textbf{89.40\%} & \textbf{78.40\%} &11 & \textbf{85.14\%} (↑10.28\%) \\ \cline{2-15} 
                    & GaitGL\cite{lin2021gait}            & 72.10\%          & 84.40\%          & \textbf{88.90\%}          & 85.00\%          & 79.50\%          & 72.30\%          & 79.70\%          & 81.80\%          & 84.80\%          & 79.20\%          & 64.50\%         &/ & 79.29\% \\ 
                    & GaitGL+PN(ours)   & \textbf{83.70\%} & \underline{89.60\%}          & \underline{88.10\%}          & \underline{88.16\%}          & \underline{85.52\%} & \textbf{87.95\%} & \underline{88.96\%} & \textbf{89.70\%}          & \underline{89.40\%}          & \underline{86.90\%}          &\underline{77.40\%} &10 & \underline{86.57\%} (↑7.28\%) \\
                    & GaitGL+PN+RMP(ours)   & \underline{82.50\%} & \textbf{89.70\%} & 88.60\%          & \textbf{90.61\%} & \textbf{88.23\%} & \underline{87.04\%} & \textbf{89.37\%} & \underline{87.80\%} & \textbf{89.80\%} & \textbf{88.80\%} & \textbf{77.60\%} &10 & \textbf{88.14\%} (↑8.85\%)\\ \hline
\end{tabular}}
\vspace{-0.3cm}
\label{tab:exp_result}
\end{table*}
%\textbf{CASIA-B} 
Table~\ref{tab:exp_result} shows the results of our proposed method and its comparison with three SOTA silhouette-based works (GaitSet, GaitPart and GaitGL) on the CASIA-B dataset. The table also shows the accuracy comparison for different view angles and walking conditions. In the table, our approaches with and without the RMP module are listed as Baseline + PN (PointNet), and Baseline + PN (PointNet) + RMP, respectively. All the results are averaged on 11 gallery views and identical views are excluded. As can be seen, Baseline+ PN + RMP outperforms all the silhouette and CNN-based baselines for BG and CL scenarios. For the CL scenario, the highest improvement is obtained and the performances of GaitSet, GaitPart and GaitGL are improved by 12.12\%, 10.28\% and 8.85\%, respectively. This result validates and is in agreement with our observation in Sec.~\ref{sec:motivation} that skeleton features contribute greatly in situations where an individual's body shape changes significantly. For NM, the performances are very comparable. While the performance of GaitSet is improved by 1.32\%, the mean accuracy compared to GaitPart and GaitGL is slightly lower (0.06\% and 0.02\% lower, respectively.)   
%and it improves the performance of the corresponding silhouettes- and CNN-based baseline for most walking scenarios in terms of mean accuracy over different view angles. 

Based on these results, we can draw the following conclusions: (i) the features extracted by the point cloud analysis model improve the performance of all the silhouette- and CNN-based baselines with varying degrees for two of the three walking scenarios, namely for the more challenging BG and CL scenarios; (ii) the more different the contours of the query and gallery images are, the higher the improvement is. 
%improvement is positively correlated with how different contours of the query and gallery images are. 
All gallery images in the CASIA-B dataset belong to the normal walking scenario. Thus, it is expected to have comparable performance from different approaches for the NM scenario. For all the baselines, our approach provides the most improvement for the wearing coat (CL) scenario, since the contour of a person wearing a coat is more different than that of a normal case, as explained in Sec.~\ref{sec:motivation}; (iii) features from the point cloud analysis model not only alleviate the problem of CNNs
overfitting to person contours but also increase the robustness to different view angles. Our method improves the accuracy for more than half of the view angles in most cases, indicating that the point cloud analysis model can learn general motion features from human key points under different view angles. The swing motions of arms, and stride length also contribute to gait features, and silhouette-based approaches have more difficulty with view angles of 0$^{\circ}$, 90$^{\circ}$ and 180$^{\circ}$ \cite{chao2021gaitset}, since these features are harder to capture from silhouettes obtained from those angles. Since sequence of skeleton key points can provide movement information for each point, our approaches provide the top 2 performances for all walking scenarios and boost the performance of all silhouette-based baselines for angles 0$^{\circ}$ and 90$^{\circ}$. For angle 180$^{\circ}$, our approaches provide the top 2 performances for BG and CL scenarios and boost the performance of all silhouette-based baselines. For NM scenario and for angle 180$^{\circ}$, our approaches provide either the best or second best performance.
%In addition, most accuracy for 0$^{\circ}$, 90$^{\circ}$ and 180$^{\circ}$ is improved and increased more than other view angles, the possible reason is that the features parallel to the walking direction, such as stride and speed, which is harder to observe at 0$^{\circ}$ and 180$^{\circ}$, but key points can provide facial information, such as eye distance and ear distance, which is difficult to be extracted in silhouettes. The vertical feature that is harder to be observed in the parallel direction, such as the swing right and left motion of arms or body, but key points can provide movement information for each point. That phenomenon indicates that skeleton can provide more detailed information at certain view angles where some features are easily missing. 
\subsection{Experiments on the OUMVLP dataset}
We also performed experiments on the OUMVLP dataset~\cite{takemura2018multi, an2020performance}, which contains gait videos of 10,307 people. Due to computational reasons, we split the dataset into eight folds and performed evaluation on three folds. For each fold, 644 people are used for training, and a different set of 644 people are used for testing. We pick up points from randomly picked 21 consecutive frames as input. In this experiment, GaitSet+PointNet provides a slight performance increase (0.55\%) for one fold, and slightly lower performance (0.4-1\%) for others. This can be explained by the differences between the CASIA-B and OUMVLP datasets. First, in the CASIA-B dataset, for each sequence there are more than 80 frames, on average, with key point information. In contrast, sequences in the OUMVLP-pose contain an average of 25 frames. Thus, there is less information provided in the OUMVLP dataset to capture and learn the key point pattern, which in turn, limits the contribution of PointNet.

The second difference is the number of available key points per frame. There are 17 and 18 key points for every frame in the CASIA-B and OUMVLP datasets, respectively. The additional point is the neck point in the OUMVLP dataset. We performed an experiment by removing the neck point to replicate the setting in CASIA-B. The performance of GaitSet+PointNet not only increased compared to using 18 points, but it also improved the performance of GaitSet by 0.19\%. 
One possible reason is that the neck point does not provide useful information for the walking pattern and can be distracting. 
\section{Ablation Studies and Analysis}
\label{sec:Ablation}
To verify the effectiveness of each component of our proposed method, ablation studies are carried out on the CASIA-B dataset. The experimental settings are the same as in Sec.~\ref{sec:exp}, 
%\ref{sssec: settings} 
and all the results in this section are rank-1 accuracy values averaged over all views excluding the identical-view cases.
\subsection{Analysis of Different Point-based Methods}
\label{ssec:Abla_pointnet}
In our approach we employ PointNet, to extract human key point features, for the reasons detailed in Sec.~\ref{ssec:GaitPointplusRMP}.
%we employ PointNet to extract the feature of human key points, 
To analyse the performance of other milestone point-based methods, we compare PointNet with DGCNN~\cite{wang2019dynamic} and GDANet~\cite{xu2021learning} without the RMP module. The results in Table.~\ref{tab:diff_pointbase} show that, in most cases, DGCNN and GDANet provide further increase over using PointNet. 
%incorporate with all baseline have better performance than PointNet. 
This is to be expected, since DGCNN and GDANet perform better than PointNet in 3D point cloud processing task in general, showing their ability
%accuracy is higher than PointNet in the field of 3D point cloud processing, that demonstrates that their ability 
to better extract point cloud features. Yet, the complexity of DGCNN and GDANet is much higher than PointNet as shown in Table.~\ref{tab: comp_point}, where the `Forward/backward pass size' and 'params size' are the memory needed to evaluate/train the model and save the model, respectively, `Total Params' is the total number of parameters contained in the model, and `Estimated Total Size' is the rough sum of Forward/backward pass size and params size. As can be seen, all parameter sizes for DGCNN and GDANet are much higher than those of PointNet. The memory requirements are also listed for the silhouette-based baselines in Table.~\ref{tab: comp_point} showing that the complexity of these silhouette-based methods is already high. Thus, it is more favorable to have the human key point processing branch as lightweight as possible.
%The branch that process the human key points is expected to be as small as possible.
PointNet is only composed of several fully connected layers, making it faster and lighter weight, which is more in line with our goals and requirements listed in Sec.~\ref{ssec:GaitPointplusRMP}.
\begin{table*}[ht!]
\centering
\caption{Comparison of using PointNet, DGCNN and GDANet (without RMP) as the skeleton processing model}
\begin{tabular}{|c|c|c|c|c|c|}
\hline
Silhouettes Processing Model & Skeleton Processing Model & NM               & BG               & CL               & mean             \\ \hline
\multirow{3}{*}{GaitSet}     & PointNet                  & 95.61\%          & 91.57\%          & 82.69\%          & 89.96\%          \\ \cline{2-6} 
                             & DGCNN                     & 95.91\%(↑0.3\%)  & 91.37\%(↓0.2\%)  & 85.11\%(↑2.42\%) & 90.8\%(↑0.84\%)  \\ \cline{2-6} 
                             & GDANet                    & 96.04\%(↑0.43\%) & 91.59\%(↑0.02\%) & 85.99\%(↑0.3\%)  & 91.21\%(↑1.25\%) \\ \hline \hline
\multirow{3}{*}{GaitPart}    & PointNet                  & 94.37\%          & 89.59\%          & 82.03\%          & 88.66\%          \\ \cline{2-6} 
                             & DGCNN                     & 94.38\%(↑0.01\%) & 90.7\%(↑1.11\%)  & 84.27\%(↑2.24\%) & 89.78\%(↑1.12\%) \\ \cline{2-6} 
                             & GDANet                    & 94.14\%(↓0.23\%) & 90.07\%(↑0.48\%) & 85.03\%(↑3.00\%) & 89.75\%(↑1.09\%) \\ \hline \hline
\multirow{3}{*}{GaitGL}      & PointNet                  & 95.70\%          & 93.49\%          & 86.57\%          & 91.92\%          \\ \cline{2-6} 
                             & DGCNN                     & 95.58\%(↓0.12\%) & 93.66\%(↑0.17\%) & 87.26\%(↑0.69\%) & 92.17\%(↑0.25\%) \\ \cline{2-6} 
                             & GDANet                    & 96.05\%(↑0.35\%) & 93.63\%(↑0.14\%) & 87.63\%(↑1.06\%) & 92.44\%(↑0.52\%) \\ \hline
\end{tabular}
\label{tab:diff_pointbase}
\end{table*}

\begin{table}[ht!]
\caption{Comparison of PointNet, DGCNN, GDANet, GaitSet, GaitPart and GaitGL in terms of size and number of parameters.}
\vspace{-0.15cm}
\centering
%\rowcolors{2}{yellow!100}{yellow!100}
\resizebox{0.95\linewidth}{!}{
\begin{tabular}{|c|c|c|c|c|}
\hline
Model    & \begin{tabular}[c]{@{}c@{}}Forward/backward \\ pass size (MB)\end{tabular}  & \begin{tabular}[c]{@{}c@{}}Total \\ Params\end{tabular} & \begin{tabular}[c]{@{}c@{}}params \\ Size (MB)\end{tabular} & \begin{tabular}[c]{@{}c@{}}Estimated Total \\ Size (MB)\end{tabular}  \\ \hline
PointNet & 2541.75                                              & 800768                            & 3.2                                   & 2546.52                                        \\ \hline
DGCNN    & 23531.36                                             & 1275008                           & 5.1                                   & 23538.02                                       \\ \hline
GDANet   & 34427.64                                             & 1851648                           & 3.7                                   & 34436.6                                        \\ \hline   \hline
% \rowcolor{yellow}
GaitSet  & 19377.68                                             & 2594592                           & 9.27                                  & 19378.4                                        \\ \hline
% \rowcolor{yellow}
GaitPart & 19645.04                                             & 975008                            & 1.8                                   & 19746.0                                          \\ \hline
% \rowcolor{yellow}
GaitGL   & 19894.24                                             & 9945441                           & 39.78                                 & 20617.28                                       \\ \hline
\end{tabular}}
\label{tab: comp_point}
\end{table}

\subsection{Analysis of the RMP module}
\label{ssec:analysis_RMP}
To validate the effectiveness of the RMP module, we perform experiments to compare the accuracy of three baselines (GaitSet, GaitPart and GaitGL) used together with three point-based methods (PointNet, DGCNN and GDANet) with and without the RMP Module. The results presented in Table~\ref{tab:rmp_effect} show the effectiveness of the RMP module when used with different point-based methods and various silhouette-based baselines. It can be seen that the mean accuracy of PointNet and DGCNN is improved with the help of RMP for all three baselines, and the improvement margin for PointNet is more significant compared to DGCNN. The impact of RMP over GDANet is not that obvious or significant. This can be explained by the fact that GDANet and DGCNN already learn many neighbor points' features, meaning their ability to learn point cloud features is already higher to begin with.  Hence, the additional contribution of recycling discarded points is limited. The performance of DGCNN by itself is also better than PointNet, explaining why the impact of RMP on DGCNN is less obvious than that on PointNet. It should also be noted that, although the performance of PointNet by itself in Table~\ref{tab:diff_pointbase} is lower than DGCNN and GDANet, PoitNet can achieve comparable and sometimes higher performance with the contribution of the RMP module. As shown in Table~\ref{tab:rmp_effect}, PointNet+RMP outperforms DGCNN+RMP when used together with GaitPart and GaitGL. PointNet+RMP also outperforms GDANet+RMP when used together with GaitGL. The maximum accuracy difference between PointNet+RMP and GDANet+RMP is only 0.44\%. It should also be mentioned that, since RMP is only employed during training, this performance improvement comes with only a slight increase in the training time without affecting the inference time. Therefore, PointNet incorporating the RMP module can get comparable or better results than the other two point-based methods, by incurring less computational cost at the same time.

\begin{table*}[]
\caption{Analysis of performance with and without RMP module on three silhouette-based baselines with three human key point processing models }
\begin{tabular}{|c|c|c|c|c|c|}
\hline
Silhouettes Processing Model & Skeleton Processing Model          & NM               & BG               & CL               & mean             \\ \hline
                             & PointNet                           & 95.61\%          & 91.57\%          & 82.69\%          & 89.96\%          \\ \cline{2-6} 
                             & PointNet+RMP                       & 96.34\%(↑0.73\%) & 91.54\%(↓0.03\%) & 84.27\%(↑1.58\%) & 90.72\%(↑0.76\%) \\ \cline{2-6} 
                             & DGCNN                              & 95.91\%          & 91.37\%          & 85.11\%          & 90.80\%          \\ \cline{2-6} 
                             & DGCNN+RMP                          & 95.15\%(↓0.76\%) & 91.07\%(↓0.3\%)  & 86.22\%(↑1.11\%) & 90.81\%(↑0.01\%) \\ \cline{2-6} 
                             & GDANet                             & 96.04\%          & 91.59\%          & 85.99\%          & 91.21\%          \\ \cline{2-6} 
\multirow{-6}{*}{GaitSet}    & GDANet+RMP                         & 95.24\%(↓0.8\%)  & 91.65\%(↑0.06\%) & 86.6\%(↑0.61\%)  & 91.16\%(↓0.05\%) \\ \hline \hline
                             & PointNet                           & 94.37\%          & 89.59\%          & 82.03\%          & 88.66\%          \\ \cline{2-6} 
                             & PointNet+RMP                       & 95\%(↑0.63\%)    & 90.95\%(↑1.36\%) & 85.14\%(↑3.11\%) & 90.36\%(↑1.7\%)  \\ \cline{2-6} 
                             & DGCNN                              & 94.38\%          & 90.70\%          & 84.27\%          & 89.78\%          \\ \cline{2-6} 
                             & DGCNN+RMP                          & 94.28\%(↓0.1\%)  & 90.59\%(↓0.11\%) & 84.86\%(↓0.59\%) & 89.91\%(↑0.13\%) \\ \cline{2-6} 
                             & GDANet                             & 94.14\%          & 90.07\%          & 85.03\%          & 89.75\%          \\ \cline{2-6} 
\multirow{-6}{*}{GaitPart}   & GDANet+RMP                         & 94.88\%(↑0.74\%) & 91.28\%(↑1.21\%) & 84.98\%(↓0.05\%) & 90.38\%(↑0.63\%) \\ \hline \hline
                             & PointNet                           & 95.70\%          & 93.49\%          & 86.57\%          & 91.92\%          \\ \cline{2-6} 
                             & PointNet+RMP                       & 95.88\%(↑0.18\%) & 93.59\%(↑0.1\%)  & 88.14\%(↑1.57\%) & 92.54\%(↑0.62\%) \\ \cline{2-6} 
                             & DGCNN                              & 95.58\%          & 93.66\%          & 87.26\%          & 92.17\%          \\ \cline{2-6} 
                             & DGCNN+RMP                          & 95.65\%(↑0.07\%) & 93.56\%(↓0.1\%)  & 87.91\%(↑0.65\%) & 92.37\%(↑0.2\%)  \\ \cline{2-6} 
                             & GDANet                             & 96.05\%          & 93.63\%          & 87.63\%          & 92.44\%          \\ \cline{2-6} 
\multirow{-6}{*}{GaitGL}     & GDANet+RMP & 95.85\%(↓0.2\%)  & 93.84\%(↑0.21\%) & 86.85\%(↓0.78\%) & 92.18\%(↓0.26\%) \\ \hline
\end{tabular}
\label{tab:rmp_effect}
\end{table*}

\subsection{Analysis of Different Loss Components}
In Eq.~(\ref{eq:final_loss}),the overall loss is composed of three triplet losses and the RMP loss. The effectiveness and the role of the RMP loss are shown in Tables~\ref{tab:exp_result} and \ref{tab:rmp_effect}, and discussed in Sec.~\ref{ssec:analysis_RMP}. We also performed experiments with one of the silhouette-based baselines, namely GaitGL, to study the role of different triplet losses on the performance. The results for the most challenging scenario of `walking while wearing a coat (CL)' as well as for average accuracy of all three classes are presented in Table~\ref{tab:ablation_loss}. As can be seen, the model incorporating all the loss terms provides the best performance. As for the individual triplet loss terms, the triplet loss for gait features ($L_{g}\__{tp}$) has the greatest impact on the result. Without $L_{g}\__{tp}$, the average accuracy drops by 15.71\%. The triplet losses for permutation invariant features ($L_{p}\__{tp}$) and convolutional features ($L_{c}\__{tp}$) play auxiliary roles. They are used to help human key point processing branch and silhouettes processing branch, respectively, so that the overall network could achieve a better performance. Without $L_{p}\__{tp}$ and $L_{c}\__{tp}$, the average accuracy drops by 0.66\% and 0.71\%, respectively.

\begin{table}[]
\caption{Analysis of role of different loss terms on performance}
\vspace{-0.15cm}
\resizebox{1\linewidth}{!}{
\begin{tabular}{|c|c|c|c|}
\hline
Model                                                   & Loss             & CL        & Mean      \\ \hline
                                & w all loss terms & 88.14\%   & 92.54\%   \\ \cline{2-4} 
                               & w/o Lc\_tp       & 86.04\%(↓2.1\%) & 91.83\%(↓0.71\%) \\ \cline{2-4} 
                               & w/o Lp\_tp       & 85.63\%(↓2.51\%) & 91.88\%(↓0.66\%) \\ \cline{2-4} 
\multirow{-4}{*}{GaitGL+PN+RMP} & w/o Lg\_tp       & 69.58\%(↓18.56\%) & 76.83\%(↓15.71\%) \\ \hline
\end{tabular}}
\label{tab:ablation_loss}
\end{table}

\subsection{Analysis of Utilization of Individual Key Points}
\begin{figure}[t!]
\centering
\vspace{-0.2cm}
\includegraphics[width=0.6\linewidth, height=0.7\linewidth]{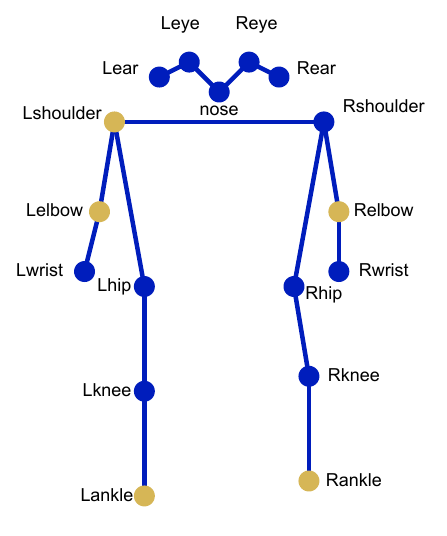}
\vspace{-0.4cm}
\caption{\small{17 human key points on a skeleton. The yellow joints are the top five most used points kept after first max-pooling.}}
\label{fig:keypoint}
\vspace{-0.2cm}
\end{figure}

We analyze the utilization of every human key point. There are 17 key points, shown on the skeleton model in Fig.~\ref{fig:keypoint}, which are $\{$nose, Reye, Leye, Rear, Lear, Rshoulder, Relbow, Rwrist, Lshoulder, Lelbow, Lwrist, Rhip, Rknee, Rankle, Lhip, Lknee and Lankle $\}$, where R and L at the beginning of words denote Right and Left, respectively. 
%For each person there are 60 frames with 17 points for each frame, resulting in $60\times 17=1020$ points in total. 
Table.~\ref{tab:point_util} shows the percentage of kept points, belonging to a body part, after one max-pooling. For instance, only $0.0698\times K_n$ of the kept points are hip points, where $K_n$ is the total number of points kept after one max-pooling (As shown in Fig.~\ref{fig:total_point_used}, $K_n$ is around 375.). For all different models, ankle and elbow points are the ones, which are most commonly preserved after one max-pooling. Although, traditional max-pooling discards a lot of the hip points, e.g., they may still contain useful information, which we aim to utilize via recycling through an RMP module. 
%In all experiment results, ankles and elbows points are the most used, nose and hips points are used relatively sparingly, but these less utilized points may contain useful information. Therefore, we use RMP to collect more points' features.   

\begin{table*}[h!]
\centering
\caption{Point utilization percentage for each body part after the first max-pooling operation}
\begin{tabular}{|c|c|c|c|c|c|c|c|c|c|}
\hline
Model             & nose   & eyes    & ears    & shoulders & elbows           & wrists  & hips   & knees   & ankles           \\ \hline
GaitSet+PointNet  & 2.96\% & 10.82\% & 10.31\% & 13.76\%   & \textbf{14.47\%} & 12.24\% & 6.98\% & 11.68\% & \textbf{16.78\%} \\ \hline
GaitPart+PointNet & 2.86\% & 9.80\%  & 9.19\%  & 13.60\%   & \textbf{16.07\%} & 12.89\% & 6.98\% & 11.81\% & \textbf{16.81\%} \\ \hline
GaitGL+PointNet   & 2.68\% & 10.51\% & 10.24\% & 12.88\%   & \textbf{14.98\%} & 13.05\% & 7.24\% & 11.58\% & \textbf{16.83\%} \\ \hline
\end{tabular}
\label{tab:point_util}
\end{table*}

\section{Conclusion}
We have proposed a new gait recognition network, referred to as the GaitPoint+, which models skeleton key points as a 3D point cloud, and employs a computational complexity-conscious 3D point processing approach to extract skeleton features, which are then combined with silhouette features for improved accuracy. We have also performed a detailed analysis of the utilization of every human key point after the use of traditional max-pooling, and have shown that while elbow and ankle points are used most commonly, many useful points are discarded by the max-pooling operation. To address this issue, we have used a Recycling Max-Pooling module, to recycle some of these discarded points during skeleton point processing, and achieved further improvement. %the number of used points before training and after training, and summarized that utilization of points may have an impact on results. Therefore, we further employ RMP module to help recycle discarded key points. 
Our proposed approach improves the performance of all the silhouette- and CNN-based baselines with varying degrees for the more challenging scenarios of carrying a bag and wearing a coat. This shows the generalizability of our approach. It has also been observed that the more different the contours of the query and gallery images are, the higher the achieved improvement is. We have also shown that 
%improvement is positively correlated with how different contours of the query and gallery images are. 
features from the point cloud analysis model not only alleviate the problem of CNNs
overfitting to person contours but also increase the robustness to different view angles.
%Our method increases the robustness against the variations in appearance due to different viewing angles or clothing or carried items, and improves the performance of different silhouettes- and CNN-based benchmarks, which shows the generalizability of our method. 
This is one of the first works that analyses the utilization of different skeleton key points for gait recognition. We have shown that the contribution of certain points is larger than others. We plan to explore the effects of different key points further as our future work.
% By incorporating features from human pose key points, via point cloud analysis, with silhouettes features to address the limitations of the only silhouettes- and CNN-based methods. We then analysed the 

%{\appendices
%\section*{Proof of the First Zonklar Equation}
%Appendix one text goes here.
% You can choose not to have a title for an appendix if you want by leaving the argument blank
%\section*{Proof of the Second Zonklar Equation}
%Appendix two text goes here.}

 % argument is your BibTeX string definitions and bibliography database(s)
%\bibliography{IEEEabrv,../bib/paper}
%

%\begin{thebibliography}{1}
%\bibliographystyle{IEEEtran}
\bibliographystyle{IEEEtran}
\bibliography{main}

% \begin{IEEEbiography}[{\includegraphics[width=1in,height=1.25in,clip,keepaspectratio]{figures/bio/burak_gray.jpg}}]{Burak Kakillioglu}
% bio here
% \end{IEEEbiography}

\begin{IEEEbiography}[{\includegraphics[width=1in,height=1.25in,clip,keepaspectratio]{image/bio/huantao-gray.pdf}}]{Huantao Ren}
received the B.S. degree in electronic information engineering from Hangzhou Dianzi University, Hangzhou, China in 2019 and the M.S. degree in electrical engineering from Syracuse University, Syracuse, NY, USA in 2021. She is currently pursuing the Ph.D. degree in the Department of Electrical Engineering and Computer Science at Syracuse University. Her research interests include gait recognition and point cloud segmentation and classification.
\end{IEEEbiography}

\begin{IEEEbiography}[{\includegraphics[width=1in,height=1.25in,clip,keepaspectratio]{image/bio/jiajing.pdf}}]{Jiajing Chen}
received the B.S. degree in mechanical engineering from Wuhan Institute of Technology, Wuhan, China in 2017 and the M.S. degree in mechanical engineering from Syracuse University, Syracuse, NY, USA in 2019. He is currently pursuing the Ph.D. degree in the Department of Electrical Engineering and Computer Science at Syracuse University. His research interests include point cloud segmentation, weakly supervised object detection and few-shot learning.
\end{IEEEbiography}

\begin{IEEEbiography}[{\includegraphics[width=1in,height=1.25in,clip,keepaspectratio]{image/bio/senem-gray.pdf}}]{Senem Velipasalar}
(M'04--SM'14) received the B.S. degree in electrical and electronic engineering from Bogazici University, Istanbul, Turkey, in 1999, the M.S. degree in electrical sciences and computer engineering from Brown University, Providence, RI, USA, in 2001, and the M.A. and Ph.D. degrees in electrical engineering from Princeton University, Princeton, NJ, USA, in 2004 and 2007, respectively. From 2001 to 2005, she was with the Exploratory Computer Vision Group, IBM T. J. Watson Research Center, NY, USA. From 2007 to 2011, she was an Assistant Professor with the Department of Electrical Engineering, University of Nebraska-Lincoln. She is currently a Professor in the Department of Electrical Engineering and Computer Science at Syracuse University.

The focus of her research has been on mobile camera applications, wireless embedded smart cameras, applications of machine learning, multi-camera tracking and surveillance systems, and automatic event detection from videos. Dr. Velipasalar received a Faculty Early Career Development Award (CAREER) from the National Science Foundation in 2011. She is the recipient of the 2014 Excellence in Graduate Education Faculty Recognition Award. She is the coauthor of the paper, which received the 2017 IEEE Green Communications and Computing Technical Committee Best Journal Paper Award. She received the Best Student Paper Award at the IEEE International Conference on Multimedia and Expo in 2006. She is the recipient of the EPSCoR First Award, IBM Patent Application Award, and Princeton and Brown University Graduate Fellowships. She is a member of the Editorial Board of the IEEE Transactions on Image Processing and Springer Journal of Signal Processing Systems.
\end{IEEEbiography}
%\newpage

%\bf{If you will not include a photo:}\vspace{-33pt}
%\begin{IEEEbiographynophoto}{John Doe}
%Use $\backslash${\tt{begin\{IEEEbiographynophoto\}}} and the author name as the argument followed %by the biography text.
%\end{IEEEbiographynophoto}

\vfill

\end{document}